%% file: root.tex
\documentclass[journal,twoside]{IEEEtran}
\usepackage{algpseudocode}
\usepackage{algorithm}
\usepackage{amsmath}
\usepackage{amssymb}
\usepackage{mathtools}
\usepackage{balance}
\usepackage{booktabs}
\usepackage[noadjust]{cite}
\usepackage{graphicx}
\usepackage{siunitx}
\usepackage[caption=false,font=footnotesize,subrefformat=parens,labelformat=parens]{subfig}
\usepackage[absolute,overlay,showboxes]{textpos}
\usepackage{booktabs}
\usepackage{makecell}
\usepackage{pgfplots}
\usepackage{flushend}

\graphicspath{{./img/}}
\pgfplotsset{compat=1.18}
\newcolumntype{R}{>{$}r<{$}}
\TPMargin*{3pt}

\let\vec\boldsymbol
\newcommand{\mlpgap}{\text{MLP}_\text{gap}}
\newcommand{\mlpacc}{\text{MLP}_\text{acc}}
\newcommand{\veh}{\nu}

\newcommand\copyrighttext{
	\footnotesize
	\noindent
	This paper has been accepted for publication in the IEEE Intelligent Transportation	Systems Magazine (ITSM). This is the accepted version of the paper, which has not been fully edited and the layout may differ from the original publication. A CC BY 4.0 version is available from https://doi.org/10.18725/OPARU-59080}%
\newcommand\copyrightnotice{%
	\textblockcolour{white}%
	\begin{textblock*}{7.1in}(0.7in,0.2in)
		\copyrighttext
	\end{textblock*}
}

\begin{document}

\title{Real-World Evaluation of two Cooperative \linebreak Intersection Management Approaches}

\author{
    Marvin~Klimke$^{*,1,2}$,
	Max~Bastian~Mertens$^{*,1}$,
	Benjamin~V\"olz$^{2}$, and
	Michael~Buchholz$^{1}$%
\thanks{*M.~Klimke and M.~B.~Mertens are both first authors with equal contribution. Names ordered alphabetically. Corresponding author: Max B. Mertens (e-mail: max.mertens@uni-ulm.de).}%
\thanks{$^{1}$Institute of Measurement, Control and Microtechnology, Ulm University, D-89081 Ulm, Germany.}%
\thanks{$^{2}$Robert Bosch GmbH, Corporate Research, D-71272 Renningen, Germany.}%
\thanks{Parts of this work were financially supported by the Federal Ministry for Economic Affairs and Climate Action of Germany within the program ``Highly and Fully Automated Driving in Demanding Driving Situations'' (project LUKAS, grant numbers 19A20004A and 19A20004F).}%
\thanks{Parts of this research have been conducted as part of the PoDIUM project, which is funded by the European Union under grant agreement No. 101069547. Views and opinions expressed are however those of the authors only and do not necessarily reflect those of the European Union or European Commission. Neither the European Union nor the granting authority can be held responsible for them.}%
}

\markboth{Real-World Evaluation of two Cooperative Intersection Management Approaches}{Klimke \lowercase{and} Mertens \lowercase{\textit{et~al.}}}

\maketitle
\copyrightnotice

\begin{abstract}
Cooperative maneuver planning promises to significantly improve traffic efficiency at unsignalized intersections by leveraging connected automated vehicles.
Previous works on this topic have been mostly developed for completely automated traffic in a simple simulated environment.
In contrast, our previously introduced planning approaches are specifically designed to handle real-world mixed traffic.
The two methods are based on multi-scenario prediction and graph-based reinforcement learning, respectively.
This is the first study to perform evaluations in a novel mixed traffic simulation framework as well as real-world drives with prototype connected automated vehicles in public traffic.
The simulation features the same connected automated driving software stack as deployed on one of the automated vehicles.
Our quantitative evaluations show that cooperative maneuver planning achieves a substantial reduction in crossing times and the number of stops.
In a realistic environment with few automated vehicles, there are noticeable efficiency gains with only slightly increasing criticality metrics.
\end{abstract}


\section{Introduction}
\label{sec:intro}

\IEEEPARstart{U}{rban} traffic is prone to inefficiencies and disturbances due to the ever-increasing volume of traffic~\cite{inrix2024scorecard}.
This manifests, e.g., at smaller intersections, where static priority rules are the prevailing method for coordination of vehicles.
Connected automated driving opens up new opportunities to improve urban traffic efficiency by leveraging communication links between vehicles and possibly infrastructure systems.
Moreover, edge computing resources are becoming available in urban areas that enable, e.g., populating and maintaining of a collective environment model (EM) of an area of interest.
This setup allows connected automated vehicles (CAVs) to improve their local planning algorithms by incorporating data from the server-side EM, as it was shown in~\cite{buchholz2021handling} at a suburban three-way intersection in Ulm-Lehr, Germany.

\begin{figure}
	\centering
	{
	\def\svgwidth{0.99\linewidth}
	\scriptsize
	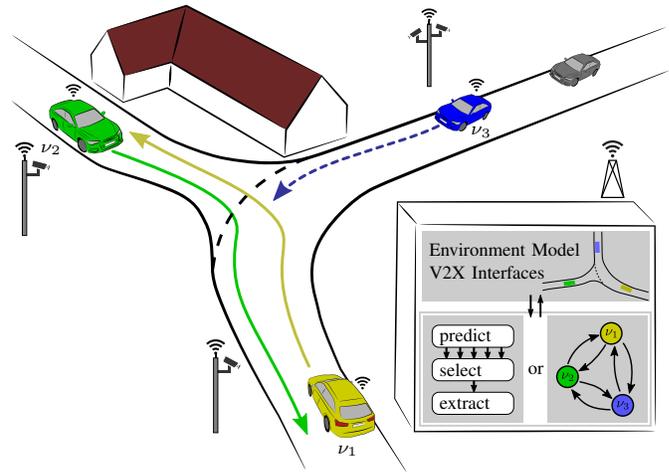
	}
	\caption{Cooperative maneuver at an unsignalized intersection.
	The CAV $\nu_3$ on the major road gives way to the turning CAV $\nu_1$ and the CAV $\nu_2$ on the subordinate road under consideration of the HDV (gray).
	We compare an optimizing planner and an RL-based planner for deployment on an edge server.}
	\label{fig:intro}
\end{figure}

Connectivity between vehicles can also be used for the coordination of maneuvers to increase traffic efficiency and safety.
In the current work, we integrate and evaluate two cooperative crossing maneuver planners, as illustrated in Fig.~\ref{fig:intro}.
The cooperative maneuvers are planned on a centralized edge server, which instructs CAVs to explicitly deviate from priority rules to reach peak efficiency.
As urban automotive traffic will not be fully automated in the near future, mixed traffic, i.e., the simultaneous use of roads by both human-driven vehicles (HDVs) and CAVs, will be prevalent.
Therefore, any real-world cooperative planner must consider HDVs that cannot be directly influenced and behave according to the priority rules.
We compare two approaches to cooperative maneuver planning, one of which leverages a multi-scenario prediction to derive an optimal maneuver, based on~\cite{mertens2022cooperative}.
The second approach employs reinforcement learning (RL) to train a graph neural network (GNN) policy for cooperative maneuver planning~\cite{klimke2023automatic}.

Our methods are designed for urban intersections with one lane per crossing direction.
Larger intersections are beyond the scope of this paper, as they are typically managed by traffic lights and require a different cooperative planning paradigm.

To achieve realistic and comparable quantitative results, we employ the real-time simulation framework DeepSIL~\cite{strohbeck_deepsil_2021} with a state-of-the-art human driver model for HDVs.
We extended the simulation framework to support multiple CAVs, maneuver planning modules, and V2X communication in between.
Also, we integrated the trajectory planner and controller combination from~\cite{ruof_real-time_2023} as well as an advanced vehicle model from one of our real-world CAVs into the simulation.
Additionally, to the best of our knowledge, we are the first to provide an experimental evaluation of cooperative planners in public mixed traffic using real test vehicles and V2X communication.
The execution of the cooperative maneuvers in simulation and in real-world is based on the coordination protocol proposed in~\cite{mertens2021extended}.
Thus, the core contribution of the present work is threefold:
\begin{itemize}
\item Extension of our simulation framework to integrate cooperative maneuvers and realistic HDV and CAV models;
\item Statistical analysis and evaluation of our two planning approaches in fully automated and in mixed traffic;
\item Demonstration of the real-world applicability of both planners using three prototypical CAVs in public traffic.
\end{itemize}

The remainder of the paper is structured as follows:
Section~\ref{sec:sota} discusses the state of the art in classical and learning-based cooperative planning for automated driving.
Our previously introduced maneuver planning approaches are summarized and compared in Section~\ref{sec:planningmodule}.
We give an in-depth overview over our simulation framework and evaluation method in Section~\ref{sec:simappr}.
Afterwards, we discuss our simulative evaluation results (Section~\ref{sec:simeval}) and real-world experiments in public traffic (Section~\ref{sec:realeval}).
Finally, Section~\ref{sec:conclusion} summarizes the results of the article and gives an outlook on future work.

\section{Related Work}
\label{sec:sota}

The efficient coordination of CAVs at urban intersections receives strong research interest in the field of automatic intersection management~(AIM).
In the following, we give an overview on published works that are deemed most relevant to cooperative maneuver planning at unsignalized intersections.
Prior works mostly consider fully automated traffic and pure simulative evaluations, as surveyed by~\cite{zhong2021autonomous}.
Also, the evaluation of the approaches often does not consider the processing time, and many researchers do not consider a communication protocol to coordinate the traffic.
Therefore, most previous works are not suitable for real-world deployment.

\subsection{Classical Cooperative Maneuver Planning}

The earliest cooperative planning methods rely on an exhaustive vehicle ordering search~\cite{li_cooperative_2006} or a simple heuristic such as first-come, first-served~\cite{dresner_multiagent_2004,wuthishuwong_vehicle_2013}.
Later approaches try to find a near-optimal crossing order of the arriving vehicles using more sophisticated algorithms like dynamic programming~\cite{yan_autonomous_2009}, ant colonies~\cite{wu_cooperative_2012}, a control policy based on Petri Nets~\cite{ahmane_modeling_2013}, Mixed Integer Quadratic Programs~\cite{hult_miqp-based_2018}, or Monte Carlo Tree Search~\cite{kurzer_decentralized_2018}.
However, all of these works are designed for CAV-only traffic and most were evaluated in simplistic simulation environments.
Only the authors in~\cite{ahmane_modeling_2013} show the real-time capability of their algorithm in simulation and in real-world experiments.

Fewer researchers have tackled the intersection management task in the much more challenging mixed traffic setting and especially scenarios with a vast majority of HDVs, which will be prevalent in the near future.
To the best of our knowledge, all previous mixed traffic approaches need to incorporate traffic lights to support the presence of HDVs.
The first notable approaches~\cite{dresner_sharing_2007,bento_intelligent_2013} were designed for a low rate of HDVs ($<$\,10\,\%) and the gains deteriorate in an environment with a higher amount of HDVs.
Two later publications~\cite{qian_priority-based_2014,yang_isolated_2016} proposed virtual platoons for signalized intersection management, supporting mixed traffic at high HDV rates and being evaluated in a real-time traffic simulation.
However, all methods mentioned above require traffic lights to control HDVs and as such are out of our problem scope.
Also, most approaches assume that the turn direction of HDVs is known, which is generally not given in real-world traffic.
Prior works claiming HDV support without traffic lights or other means of communication with human drivers depend on strong assumptions on the HDVs' behavior in mixed traffic.
The reservation-based approach~\cite{zhong2024reservation} assigns CAVs a priority based on their lane, while expecting HDVs to follow a possibly present leading CAV closely.
The application of such an approach in public traffic where HDVs are not specifically instructed remains questionable.

Our optimization-based approach proposed in~\cite{mertens2022cooperative} is specifically designed to handle unsignalized intersections with any HDV rate and unknown HDV turn directions.
It employs a scene-consistent multi-scenario prediction and generates maneuvers between the present CAVs to optimize the expected efficiency.
A respective coordination protocol to execute the cooperative maneuvers on CAVs was previously proposed in~\cite{mertens2021extended}, making this approach applicable on real vehicles.

\subsection{Learning-based Cooperative Maneuver Planning}

Although non-learning approaches are still prevalent in AIM, machine learning is also gaining traction in connected automated driving.
In~\cite{cui2022coopernaut}, it is proposed to leverage collective perception to improve the performance of local planning algorithms.
The planning task only considers one automated vehicle without cooperative objective, though.
Due to lack of ground-truth data for training, learning-based approaches to AIM typically rely on RL.
Such a learning-based approach is proposed by~\cite{wu2019dcl-aim} that suggests to train a policy through RL to choose from a restricted action space that ensures collision-free maneuvers.
However, this approach is designed for CAV-only traffic and the evaluation relies solely on simulations.

In~\cite{quang_tran2020proximal}, the authors propose an RL approach managing intersection traversals in presence of simulated human-driven vehicles while turning maneuvers are disallowed.
A further RL framework that combines local observations with a joint reward to accommodate a cooperative objective was presented by~\cite{guo2022coordination} and is capable of handling mixed traffic, though with a limited number of cooperating vehicles in the scene.
Both works lack the application to real vehicles and a comprehensive evaluation.

In our previous works~\cite{klimke2022cooperative,klimke2022enhanced}, we have proposed a flexible graph-based scene representation and an RL training scheme for AIM in fully automated traffic.
It was shown that the model outperforms a FIFO baseline and generalizes within certain limits to intersection layouts not encountered during training.
Our RL-based cooperative planning model has been extended to mixed traffic in~\cite{klimke2023automatic}.
This approach also does not rely on the turn direction of the HDVs to be known.
In addition, the integration of the trained RL policy with a sampling-based motion planner was addressed in~\cite{klimke2023integration}, demonstrating the applicability on real vehicles.

\section{Cooperative Maneuver Planning Modules}
\label{sec:planningmodule}

In this section, the commonalities and differences of the two evaluated maneuver planning modules are presented.
In our proposed system architecture, all CAVs in the scene provide their current state and desired route in the form of a sparsely sampled polyline to the centralized cooperative planning module on the edge server.
The planning module generates a joint maneuver on behavior level, which is in turn distributed to all CAVs.

\subsection{Definition and Coordination of Cooperative Maneuvers}

For both planning modules, the maneuvers are passed to the CAVs according to the maneuver coordination protocol proposal in~\cite{mertens2021extended}.
Depending on the number of vehicles on conflicting paths, up to two maneuver waypoints (one at the intersection entry and one at the intersection exit) are passed to a given CAV.
The CAVs need to fulfill constraints defined by the maneuver waypoint as
\begin{equation}
    \mu =  (\boldsymbol{p},\,t_\mathrm{min},\,t_\mathrm{max},\,\nu^\uparrow,\,\nu^\downarrow).
    \label{eq:maneuver_wp}
\end{equation}
The interval $[t_\mathrm{min},\,t_\mathrm{max}]$ denotes the admitted time window for the vehicle to cross the waypoint $\boldsymbol{p}$.
The fields $\nu^\uparrow$ and $\nu^\downarrow$ describe the preceding and following vehicle IDs, respectively.
This additional information can be leveraged by vehicle-side motion planners to improve follow trajectory planning on lead vehicles~\cite{ruof_real-time_2023}.
Each CAV remains responsible for its own motion planning and validation whether the maneuver constraints can be safely incorporated.
Note that lateral guidance is provided by lane centerlines in a common map that is available to all vehicle-side motion planners.

\subsection{Environment Model Processing}

The basis for both cooperative maneuver planning modules is the server-side EM.
This EM is the result of fusing multiple data sources, like cooperative awareness messages (CAM, \cite{etsi2014cam}) and collective perception messages (CPM, \cite{etsi2019cpm}) from connected vehicles or infrastructure perception~\cite{buchholz2021handling}.
It is considered to contain all information relevant for maneuver planning with full observation of the intersection area.
A vehicle in the EM is denoted as
\begin{equation}
    \nu_\mathrm{id} = (\boldsymbol{T},\,v,\,\mathcal{D},\,c) \in \mathrm{EM},
    \label{eq:em}
\end{equation}
where $\mathrm{id}$ denotes a unique identifier and the binary flag $c$ determines whether the vehicle is a CAV and thus controllable.
$\boldsymbol{T}$ describes the vehicle's pose on the local 2D ground plane and $v$ is the current driving speed.
CAVs share their intended route or destination $\mathcal{D}$ as a set of sparse waypoints, which is unknown for HDVs, though, which are assumed to be non-connected.

In a preprocessing step, both planning modules associate and project the vehicles in the server-side EM to the lane centerlines of an internal map format.
While connected vehicles share their routes with the planning module, the turn intention of human drivers is unknown.
Thus, both planners conservatively assume that an HDV may be in conflict with any crossing or merging lanes.
Additionally, it is assumed that HDVs do not know the route of other vehicles and that the indicator light states cannot be reliably evaluated.
The handling of this uncertainty might be improved by employing a prediction algorithm such as~\cite{strohbeck2020multiple}, which is out of scope for this work, though.
Moreover, vulnerable road users that are not lane-bound are not accounted for in the present setup.

\subsection{Optimization-based Planning Algorithm}
\label{sec:optim}

The optimization-based planner builds upon our previous works presented in~\cite{mertens2022cooperative}.
The main component is a scene-consistent prediction module that estimates the behavior of the vehicles of the current traffic scene in the immediate future.
Multiple such scenario predictions with various applicable maneuvers are performed and the best one---according to the validity and efficiency metrics defined below---is selected.

\subsubsection{Driver Model for Prediction}

\begin{figure}[t]
	\centering
	\def\svgwidth{0.95\linewidth}
	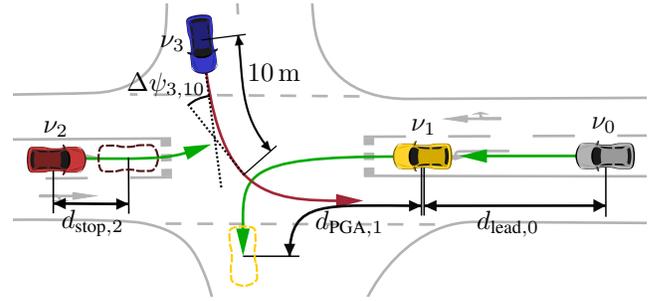
	\caption{%
		Illustration of the observation input features listed in Table~\ref{tbl:agent_observation}.
		Figure adapted from~\cite{mertens_fast_2024}.
	}
	\label{fig:agent_observation}
\end{figure}

Vital to the maneuver planning is the scenario prediction module that we previously published in~\cite{mertens_fast_2024}.
We employ a longitudinal lane-based prediction module trained on real-world traffic data.
The driver model comprises two multi-layer perceptrons (MLPs) with 2 hidden layers of size 16 each: $\mlpacc$ estimates the acceleration during the next time step.
$\mlpgap$ determines the gap acceptance, i.e., the decision whether a side road vehicle should merge or cross in a gap in main road traffic.
The driver model is evaluated jointly for all vehicles in the traffic scene and integrated over a {15\,s} horizon with a time step of {0.1\,s}.
This prediction is run for each priority assignment as described in Section~\ref{sec:prio_assignm}.
The input features are shown in Table~\ref{tbl:agent_observation} and Fig.~\ref{fig:agent_observation}.

\begin{table}[t]
	\vspace{0.2cm}
	\caption{Observation input features of the two prediction MLPs of the optimization-based planning module.
		Table adapted from~\cite{mertens_fast_2024}.}
	\label{tbl:agent_observation}
	\begin{center}
		\begin{tabular}{l l}
			\toprule
			\thead[l]{Environment observation $\vec{o}_{i}$ of $\veh_i$} & \thead[l]{Input features} \\
			\midrule
			Distance to stop line & $d_{\text{stop},i}$ \\
			Current velocity and speed limit & $v_i$, $v_{\max,i}$ \\
			Relative lane heading in $n$ meters & \makecell{$\Delta\psi_{i,-10}$, $\Delta\psi_{i,-3}$, $\Delta\psi_{i,3}$, \\ $\Delta\psi_{i,10}$, $\Delta\psi_{i,30}$, $\Delta\psi_{i,100}$} \\
			Lead vehicle distance and velocity & $d_{\text{lead},i}$, $v_{\text{lead},i}$ \\
			\toprule
			\thead[l]{Gap obs. $\vec{o}_{i,j}$ of $\veh_i$ towards $\veh_j$} & \thead[l]{Input features} \\
			\midrule
			Distance to end of conflict zone & $d_{\text{target},i}$ \\
			Velocity of this and the other vehicle & $v_i$, $v_j$ \\
			Distance of other vehicle to stop line & $d_{\text{stop},j}$ \\
			\bottomrule
		\end{tabular}
	\end{center}
\end{table}

The acceleration estimation $\mlpacc$ was trained using proximal policy optimization (PPO)~\cite{schulman_proximal_2017} in an RL closed-loop simulation environment on different intersection layouts.
The objective function encourages driving near the speed limit and penalizes collisions with prioritized vehicles to emulate natural driving behavior.
The gap acceptance model $\mlpgap$ was trained using supervised learning on real driving data extracted from the inD dataset~\cite{bock2020ind}.

\subsubsection{Priority Assignment Search}
\label{sec:prio_assignm}

Each possible maneuver is internally represented as a list $\mathcal{P}$ of priority assignments between pairs of cooperative vehicles.
Each entry is a pair of vehicles $\langle\veh_i,\veh_j\rangle$, where the first is prioritized over the second.
The planner operates cyclically in a frequency of {5\,Hz} and tries to add assignments to the previous cycle result to optimize the efficiency, by predicting the scenario for each $\mathcal{P}$.
As the future behavior of HDVs, such as gap acceptance and turning direction, is unknown and cannot be controlled, all probable actions of HDVs have to be considered.
Thus, a dependency graph analysis only considers maneuvers independent of HDV behavior as valid, largely reducing the number of feasible cooperative maneuvers in mixed traffic.

As an efficiency metric $e(\mathcal{P})$ to evaluate potential maneuvers, we use the relative velocity integrated over a prediction horizon $\left[T_\text{start},T_\text{end}\right]$ with an additional small penalty on the maneuver complexity:
\begin{align}
	\label{eq:eff}
	e(\mathcal{P}) &\coloneqq - \SI{1}{s}\cdot\left|\mathcal{P}\right| + \sum_{\veh_i\in\mathrm{EM}} \int_{T_\text{start}}^{T_\text{end}} \frac{v_i(t)}{v_{\max,i}(t)} \,\text dt,
\end{align}
where $v_{\max,i}(t)$ denotes the lane speed limit of vehicle $\veh_i$ at its position at time $t$.
The scenario prediction for each $\mathcal{P}$ is checked for freedom from collision and for correct crossing order according to the respective priority assignments.
The most efficient of the valid priority assignment lists is selected as the result $\mathcal{P}_k$ of the current planning cycle.

The number of possible priority assignments and, therefore, the simulation and search run time rises with the number of conflicting CAV pairs.
To remain real-time capable in larger scenarios, the search is limited to the first 100 priority assignments, keeping the planning time well below \SI{200}{ms}.
This limit was, however, not reached in the considered scenarios.

\subsection{RL-based Planning Algorithm}
\label{sec:rl}

The RL-based cooperative planner combines the methods presented in our previous works \cite{klimke2023automatic} and \cite{klimke2023integration}.

\subsubsection{Learning Model}
\label{ssec:learningmodel}

Due to the lack of ground-truth data for cooperative maneuvers in urban traffic, we consider the planning task a multi-agent RL problem.
The joint planning task is modeled by a single partially observable Markov decision process, since the driving intention and destination of the HDVs is not known to the planning algorithm.
Thus, the dimensionality of the observation space depends on the number of vehicles currently in the scene and may vary over time.
Moreover, the dimensionality of the action space depends on the number of cooperative vehicles present and requires a permutation equivariant mapping to the observation space.

The RL~policy is trained in a simulated traffic environment, based on the open-source simulator Highway-env~\cite{highway-env}, which has been extended for multi-agent planning.
It shall derive a joint action for all CAVs in the scene composed of longitudinal acceleration commands in continuous space.
We employ the TD3~\cite{fujimoto2018addressing} actor-critic RL algorithm to train a GNN for the cooperative planning task.

\subsubsection{Graph-based Representation and Network}
\label{ssec::graphrepresention}

\begin{figure}
	\centering
	\includegraphics[width=\linewidth]{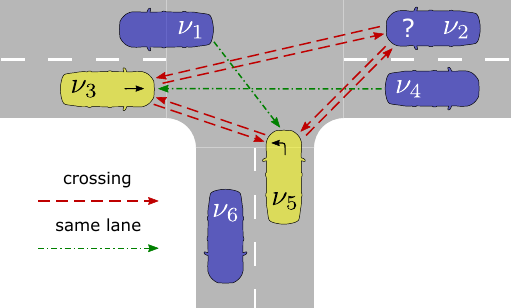}
	\caption{Graph-based input representation for mixed traffic of the RL-based planning module. A CAV's (yellow) turning intention is denoted by an arrow on its hood. Due to the unknown turning intention of the HDV $\nu_2$ (blue, denoted by '?'), it shares edges with both CAVs, although only the conflict with $\nu_5$ is inevitable.}
	\label{fig:scene_graph}
\end{figure}

The graph based scene representation used in this study is based on our proposal in~\cite{klimke2023automatic}, which proved to be suited for learning a sensible mixed traffic capable RL policy.
Thus, the current traffic scene observation is defined as a directed graph, as illustrated in Fig.~\ref{fig:scene_graph}.
The nodes of vehicles that are in conflict and thus need to be coordinated are connected by directed edges, while it is assumed that vehicles follow and stay within their lanes.
Due to the inherently unknown future behavior of HDVs, including gap acceptance and turning intention, the graph comprises additional edges for all potential conflicts, like with $\nu_2$ in the figure.
Only once the maneuver of the HDV becomes clear (typically while entering the intersection), the superfluous edges are removed and the cooperative maneuver is subsequently optimized.
Different kinds of interactions are encoded by means of edge types.
The \emph{crossing} edge type indicates a pair of vehicles that are located in front of the intersection driving on paths that intersect or merge on the intersection area.
Leader-follower relations are encoded by an edge of type \emph{same lane} pointing from the leader to the following vehicle.

In addition to the semantic structure encoded in the graph, both vertices and edges are augmented by input feature sets.
While the vertex input features contain vehicle-relevant measurements, the edge features are composed of pair-wise properties like distances.

The first part of the network architecture is comprised of MLPs that encode the vertex input features and edge input features, respectively.
At the core of the GNN, modified relational graph convolutional network layers~\cite{gangemi2018modeling} and graph attention layers~\cite{velickovic2018graph} are used for message passing in an interleaved structure.
In the actor network architecture, the resulting vertex features are mapped to a joint action by a final output MLP.
In contrast, the critic network aggregates the vertex feature vectors resulting from message passing to a single feature vector before decoding a Q-value estimate using another MLP.
The graph representation and the GNN are implemented using the PyTorch Geometric API~\cite{pyg}.

Apart from a suited network architecture, RL requires a well-designed reward function to learn a reasonable policy.
Details on the definition and tuning of the various reward components can be found in~\cite{klimke2022cooperative,klimke2023automatic}.

RL planning runs are triggered regularly With a period of \SI{2}{\second}, which facilitates the observation of the CAVs' reactions to the cooperative maneuver with the start of the next planning run.
For typical traffic scenarios at smaller intersections, the planning run takes significantly less than the period length of \SI{2}{\second}.
In unfavorable circumstances with many (non-connected) vehicles, a full planning run might take longer, which is caught by a timeout that ensures real-time capability \cite{klimke2023integration}.

\subsection{Prediction and Derivation of Cooperative Maneuvers}

For the optimization-based planner, the selected priority assignments need to be converted to maneuver constraints as in Eq.~\eqref{eq:maneuver_wp} for each individual CAV.
Each pair of conflicting CAVs shares a conflict zone, i.e., a segment along the road that may only be occupied by one of the vehicles at any time to guarantee freedom from collision, as explained above.
The predicted timestamp at which the prioritized vehicle is predicted to leave the conflict zone is used to encode the space-time maneuver constraints for entering and exiting.
The start and end points of the conflict zone along the respective vehicle routes are used as waypoints for the constraints.

The RL-based cooperative planner derives cooperative maneuvers using a built-in simulator.
After initializing the simulator based on the current state of the server-side EM, the RL~policy is queried to predict the future evolution of the scene.
Analogously to the optimization-based planner, the crossing points of the simulated trajectory and the intersection conflict zone are used to derive so-called waypoint candidates.
Only if there is a conflicting vehicle, the corresponding candidate waypoint is used as a maneuver waypoint.
Thus, only if there is a preceding vehicle, the waypoint at the intersection entry is included in the maneuver.
Otherwise, the CAV is free to enter the intersection as soon as it considers it beneficial to the maneuver, which leaves room for optimization to the CAV motion planning.

As noted above, the CAVs remain responsible to check the maneuver for local feasibility, and the maneuver planners cannot guarantee that the CAVs can execute the suggested maneuver.
However, while the high-level maneuver planning is optimized for speed, the driver models internally used in each maneuver planner strive to resemble realistic behavior, which usually is far from the physical (or even comfortable) limits at intersections.
Thus, the coordinated CAVs can usually easily comply with the required time constraints.

\section{Simulation Approach}
\label{sec:simappr}

We integrated the proposed cooperative planning approaches into an extended simulation and evaluation framework for fully automated and mixed traffic.
The cooperative planners are evaluated and assessed against two baselines.
Our primary baseline are non-cooperative (NC) CAVs that leverage shared perception data as in~\cite{buchholz2021handling}, but no routing information and adhere to priority rules.
This baseline employs the same underlying trajectory planning as is used for the cooperative approaches.
Thus, it constitutes a rather strong baseline, which facilitates a sharp evaluation of the cooperative aspect in isolation.
In addition, we present the metrics for pure HDV traffic, which is simulated using a learned behavior model.
As such, it resembles human driving behavior more closely and exhibits a human-typical agile behavior.
The focus of this paper is on the comparison of the two cooperative planning approaches at one exemplary intersection.
We refer the reader to previous publications for more extensive individual evaluations of the optimization-based \cite{mertens2022cooperative} and the RL-based \cite{klimke2023automatic} cooperative planners, respectively.

\subsection{Simulation Setup}
\label{ssec:simsetup}

The simulation framework that we used for evaluation of the planning methods builds upon the DeepSIL simulator introduced in~\cite{strohbeck_deepsil_2021}.
It is based on the ROS2 software distribution \cite{ros2} and can simulate HDVs as well as CAVs.
We extended the framework to support multiple CAVs and cooperative maneuvers.
The simulation has a base time step of {50\,ms} and runs in wall-clock time.
This means that processing delays from the cooperative maneuver planners are considered and affect the results, while communication delays are not modeled.

We use a single potent simulation machine with an AMD Ryzen 3990X CPU.
Still, running multiple instances of large parts of the CAV software stack as well as multiple deep-learning-based HDV models requires a lot of computational resources, while the maneuver planning requirements are comparatively modest.
Due to those limitations, the simulations need to be limited to eight vehicles total.
Additionally, scenarios with more than four CAVs are slowed down slightly to cope with immaturity in the ROS2 communication layer.

Human-driven vehicles are predicted using the approach in~\cite{strohbeck_deepsil_2021}.
Every {50\,ms}, the multiple trajectory prediction network is evaluated on the current traffic environment.
In each cycle, the most progressing, non-conflicting prediction is used to update the vehicle position and velocity.
The prediction is converted into a longitudinal action and the internal vehicle simulation is stepped forward along the centerline of the route.
A simple occlusion simulation was added, i.e., HDVs on the subordinate road approach slowly until they reach a distance of {2\,m} to the intersection, at which a full view into the situation is assumed.
Also, the simulated HDVs do not know the routes of the other vehicles and therefore need to assume the most conflicting route.

CAVs are simulated using a single-track model with friction and dead time, resembling one of our real-world CAVs.
The inputs to this model are computed by the trajectory planner and controller from~\cite{ruof_real-time_2023}, which is part of this CAV's software and therefore provides a realistic CAV behavior.
Like HDVs, CAVs do not know the route of other vehicles, even of other CAVs, and assume the most conflicting route.
However, the simulated CAVs do not need to approach the intersection slowly due to occlusion, because they receive the intersection EM as in~\cite{buchholz2021handling}.

The two approaches for cooperative maneuver planning are developed and integrated as alternative ROS2 software modules.
For each simulation run, only one of the modules is activated (or none in the non-cooperative case), while the remaining simulation setup stays identical.
They communicate with the simulated CAVs and server-side EM using the V2X communication protocols we proposed in \cite{buchholz2021handling,mertens2021extended}.

\subsection{Scenario Definition}
\label{ssec:scenario}

\begin{figure}
	\centering
	\includegraphics[width=\linewidth]{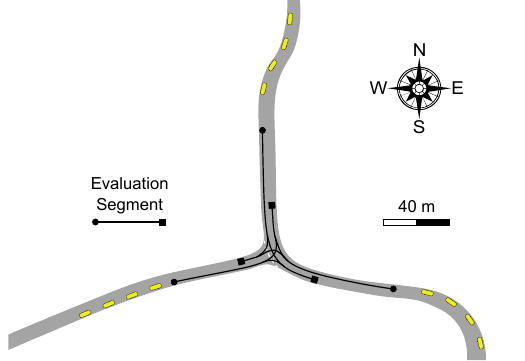}
	\caption{The initial positions of vehicles (yellow boxes) in the simulated scenarios. They are populated starting with the positions closest to the intersection. The evaluated trajectory segments are indicated by the black paths. The minor subordinate road approaches from the west.}
	\label{fig:scenarios}
\end{figure}

We define all simulated scenarios at an exemplary intersection in Ulm-Lehr, Germany, as shown in Fig.~\ref{fig:scenarios}, where also our pilot site (cf. Section~\ref{ssec:pilotsite}) is located for coherent evaluations.
It is a T-junction with a bending main road and a minor subordinate road.
The speed limit is \SI{11.11}{\meter\per\second} (\SI{40}{\kilo\meter\per\hour}) on the eastern and western part and \SI{8.33}{\meter\per\second} (\SI{30}{\kilo\meter\per\hour}) on the northern part.

\begin{figure*}
	\centering
	\subfloat[Scenario crossing order]{\includegraphics[width=0.33\textwidth]{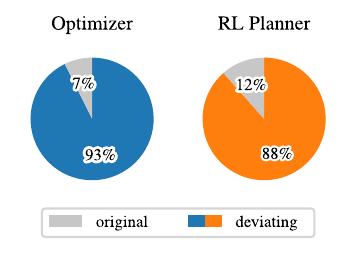}\label{sfig:all_cav_deviating}}
	\hfill
	\subfloat[Maneuver duration relative to NC~baseline]{\includegraphics[width=0.33\textwidth]{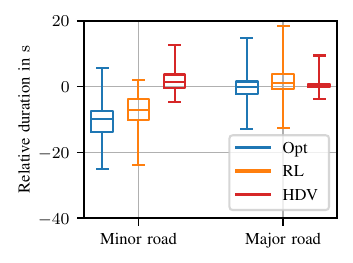}\label{sfig:all_cav_relative_duration_road_type}}
	\hfill
	\subfloat[Share of stopped vehicles]{\includegraphics[width=0.33\textwidth]{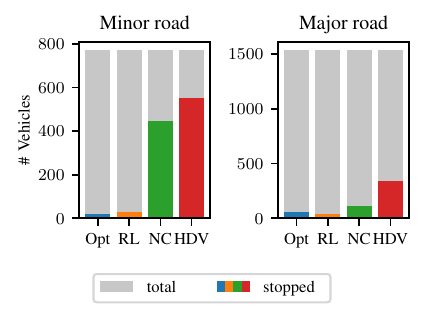}\label{sfig:all_cav_stopped_vehicles_road_type}}
	\caption{The simulative evaluation results in fully automated traffic for the optimization-based cooperative planner~(Opt), the RL-based cooperative planner~(RL), and non-cooperative~(NC) CAVs. HDV denotes pure HDV traffic. Fig.~\protect\subref*{sfig:all_cav_relative_duration_road_type} and Fig.~\protect\subref*{sfig:all_cav_stopped_vehicles_road_type} consider all scenarios in which at least one cooperative planner suggests a deviating crossing order.}
	\label{fig:sim_all_cav}
\end{figure*}

\begin{figure*}
	\centering
	\subfloat[Scenario crossing order]{\includegraphics[width=0.33\textwidth]{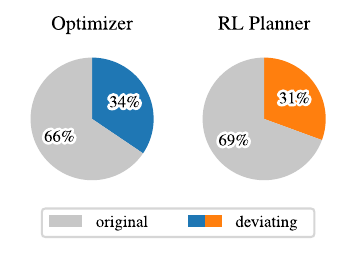}\label{sfig:mixed_deviating}}
	\hfill
	\subfloat[Maneuver duration relative to NC~baseline]{\includegraphics[width=0.33\textwidth]{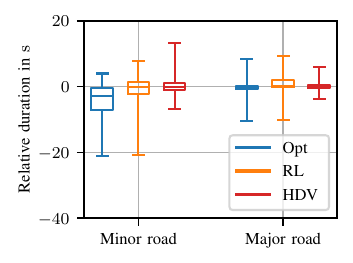}\label{sfig:mixed_relative_duration_road_type}}
	\hfill
	\subfloat[Share of stopped vehicles]{\includegraphics[width=0.33\textwidth]{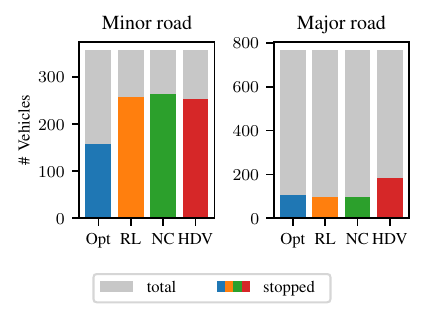}\label{sfig:mixed_stopped_vehicles_road_type}}
	\caption{The simulative evaluation results in mixed traffic under 50\,\% automation level for the optimization-based cooperative planner~(Opt), the RL-based cooperative planner~(RL), and non-cooperative~(NC) CAVs. HDV denotes pure HDV traffic. Fig.~\protect\subref*{sfig:mixed_relative_duration_road_type} and Fig.~\protect\subref*{sfig:mixed_stopped_vehicles_road_type} consider all scenarios in which at least one cooperative planner suggests a deviating crossing order.}
	\label{fig:sim_mixed}
\end{figure*}

Within this setting, we created a total set of 480 scenarios that serve as an initial configuration to the simulation.
In each scenario, a number of vehicles is spawned at the locations shown in Fig.~\ref{fig:scenarios}, where each vehicle has an initial speed of \SI{8}{\meter\per\second} and a specific route, i.e., turn direction.
The first 280 scenarios are an enumeration of all combinations of source and target directions with up to two vehicles from each direction and in which there is at least one pair of conflicting vehicles.
The remaining 200 scenarios are randomly sampled with up to four vehicles from each direction and a random route, limited to eight vehicles total as explained above.

The scenario set was simulated both in fully automated traffic and mixed traffic.
When simulating mixed traffic, the vehicles are randomly divided into {50\,\%} HDVs and {50\,\%} CAVs, where for an odd number of vehicles, the last one is chosen randomly.
Out of the 480 total scenarios, less than {3\,\%} of the simulations were not successful and resulted in a timeout or software crash (five scenarios in fully automated traffic, 14 scenarios in mixed traffic).
Those issues are unrelated to the maneuver planning and were caused by other CAV software modules and immaturity in the ROS2 communication layer. Thus, those scenarios were excluded from the evaluation in all maneuver planning configurations.

\section{Simulation Evaluation Results}
\label{sec:simeval}

The extensive simulation experiments were quantitatively evaluated in terms of efficiency metrics as well as criticality metrics, which is presented in the following section.
To ensure a fair comparison, all trajectories have been clipped to common evaluation intervals, as illustrated in Fig.~\ref{fig:scenarios}.
The following evaluations refer to trajectory segments that begin 60\,m in front of the intersection entry ($\bullet$ in Fig.~\ref{fig:scenarios}) and end 15\,m behind the intersection ({\scriptsize$\blacksquare$} in Fig~\ref{fig:scenarios}).
These intervals cover the relevant part of the intersection approach during which the CAVs synchronize onto their assigned slot for crossing the intersection.

\subsection{Effectiveness and Efficiency Analysis}
\label{ssec:quantitativeresults}

Fig.~\ref{fig:sim_all_cav} depicts the results in fully automated traffic.
Both planners are able to realize cooperative maneuvers in the vast majority of 
scenarios, as can be seen from Fig.~\subref*{sfig:all_cav_deviating}.
The ``original'' order is the one resulting from simulating without cooperation.
In general, the crossing order is not derivable from static priority rules, as gaps in main road traffic can be used.
In our specific scenario, however, there are no gaps, so the non-cooperative crossing order is unambiguous and any deviation results from an applied cooperative maneuver.

For comparability, the analyses in this section are performed over all explicitly cooperative scenarios, that is, all scenarios in which at least one cooperative planner deviates from the baseline crossing order.
For all other scenarios, the metrics do not significantly differ from the baseline.
In the cooperative scenarios, the traffic flow especially on the subordinate road is improved significantly, while the optimization-based planner provides an additional benefit over the RL-based one.
Fig.~\subref*{sfig:all_cav_relative_duration_road_type} depicts the distribution of the duration required by the vehicles to complete the maneuver relative to non-cooperative CAVs.
Per definition, NC~traffic exhibits a relative duration of \SI{0}{\second} and is thus not shown in the plot.
Almost all vehicles on the subordinate road benefit of a reduced delay due to interaction.
Some vehicles on the major road take slightly longer to complete the maneuver in the cooperative case.
This is expected since the vehicles on the major road waive their unconditional right of way when taking part in cooperative maneuvers.
Overall, the cooperation leads to a reduction in median duration, which becomes even more pronounced when comparing to pure HDV traffic.
The optimization-based planner achieves slightly higher efficiency due to planning with a lower safety distance, as discussed in Section~\ref{ssec:criticality_analysis}.

Having a vehicle stop is particularly disadvantageous in terms of both energy efficiency and passenger comfort.
As can be observed in Fig.~\subref*{sfig:all_cav_stopped_vehicles_road_type}, both cooperative planners eliminate most of the stops required on the subordinate road when driving according to priority rules.
Notably, the share of stopped vehicles on the major road does not increase but even decrease slightly, which can mainly be attributed to left-turning vehicles yielding to oncoming traffic.
Pure HDV traffic exhibits much more stops because they cannot rely on shared perception data and thus have to approach the intersection more carefully.
Meanwhile, the NC baseline only leads to a moderate decrease in stops, despite its huge perception advantage.

In mixed traffic, both cooperative planning approaches increase the traffic efficiency, as illustrated by Fig.~\ref{fig:sim_mixed}, although to a lesser extent than in fully automated traffic.
It is worth noting that a 50\,\% share of HDVs renders many cooperative maneuvers infeasible because the priority relations towards HDVs have to be retained.
Thus, achieving a notable gain in efficiency is much more challenging in mixed traffic, which is consistent with the individual planner evaluations in \cite{mertens2022cooperative,klimke2023automatic}.
We refer to these publications for more in-depth simulative and dataset-based evaluations with multi-lane intersections.

Fig.~\subref*{sfig:mixed_relative_duration_road_type} indicates that still many vehicles on the subordinate road benefit of a reduced delay due to the cooperative maneuver.
It is important to note that the simulation setup does not resemble continuous traffic but only the drive off of the initial vehicle configuration.
Unlike in real traffic, the vehicles on the subordinate road will be able to cross the intersection fluently after the prioritized traffic has left the intersection.
This circumstance leads to a favoring of the baseline maneuvers, which might have taken much longer in real continuous traffic.
Nonetheless, a reduction of stops can be observed in Fig.~\subref*{sfig:mixed_stopped_vehicles_road_type}.

The optimization-based planner clearly outperforms the RL-based planner in terms of stops on the subordinate road in mixed traffic.
This can most likely be attributed to the training of the RL being performed in continuous traffic to facilitate learning of a robust policy.
In the evaluated case of single scenarios, the learned efficiency estimation of the RL model might not be accurate.
Also, the optimization-based planner seems to favor traffic efficiency while compromising in terms of safety distance, which is examined in the following section.

\subsection{Criticality Analysis}
\label{ssec:criticality_analysis}

The gain in traffic efficiency due to cooperative maneuver planning should not be realized at the expense of safety.
Thus, we complement the performance analysis by a maneuver criticality analysis based on three metrics:
\begin{itemize}
    \item The two-dimensional time-to-collision (TTC) is a generalization of the TTC metric to intersection scenarios;
    \item The second criticality metric is defined as the deceleration required to avoid a collision (DRAC);
    \item Finally, we report the post encroachment times (PET) for the maneuvers.
\end{itemize}
We calculate the TTC using the implementation from~\cite{jiao_fast_2023}.
For the DRAC metric, the post-processing takes each simulated time step and calculates the required deceleration of each vehicle to avoid collision in case all other vehicles would suddenly decelerate with \SI{3.4}{\meter\per\second\squared}.
The PET between two conflicting vehicles is the duration between the first vehicle leaving and the second vehicle entering the shared conflict zone.

For each simulated scenario, we gather the most critical (i.e., lowest TTC and PET, highest DRAC) metrics across all vehicles and time steps.
Then, we report the median value $m_i$ across all evaluated scenarios and the percentage of critical scenarios $p_i$ ($i\in\left\{\text{TTC},\text{DRAC},\text{PET}\right\}$).
For this classification, we apply commonly accepted thresholds of \SI{1.5}{\second} for TTC, \SI{3.4}{\meter\per\second\squared} for DRAC (both taken from~\cite{huber_evaluation_2020}), and \SI{1}{\second} for PET~\cite{paul_post_2020}.%

The criticality results for fully automated traffic and for mixed traffic are depicted in Table~\ref{tab:criticality_all_cav} and Table~\ref{tab:criticality_mixed}, respectively.
Please note that we evaluated all and not only the deviating scenarios as the analysis of the criticality metrics shall only convey qualitative trends.
For this reason, there is no claim about the absolute numbers or in-depth discussion.

Both planner methods yield a higher share of critical scenarios compared to the baseline, which is expected as they are optimized for efficiency.
The RL~planner results are closer to the NC~baseline, which is most likely due to the strong focus on HDV interoperability in the reward scheme and training procedure.
Meanwhile, the optimization-based planner shows significantly increased criticality percentages for DRAC and PET, which goes along with its better performance in traffic efficiency metrics.
This is due to the fact that, internally, the planner calculates geometrically accurate conflict zones and thus assumes a PET of 0\,s to be sufficiently safe.
Increasing this safety distance to allow for more reaction time will be considered in future works.%

\begin{table}
	\caption{Criticality metrics in fully automated traffic.}
	\label{tab:criticality_all_cav}
	\centering
	\begin{tabular}{lRRRRRR}
		\toprule
		\textbf{Planner}\hspace{-0.15cm} & \frac{m_\text{TTC}}{\si{s}} & \frac{p_\text{TTC}}{\si{\percent}} & \frac{m_\text{DRAC}}{\si{ms^{-2}}} & \frac{p_\text{DRAC}}{\si{\percent}} & \frac{m_\text{PET}}{\si{s}} & \frac{p_\text{PET}}{\si{\percent}} \\
		\midrule
		\textbf{Opt} & 5.9 & 2.9 & 1.7 & 12.6 & 1.4 & 35.1 \\
		\textbf{RL}  & 4.3 & 1.7 & 1.0 &  2.5 & 1.8 &  7.3 \\
		\textbf{NC}  & 4.5 & 2.5 & 0.6 &  1.0 & 1.8 &  3.8 \\
		\bottomrule
	\end{tabular}
\end{table}

\begin{table}
	\caption{Criticality metrics in mixed traffic (50\,\% automation).}
	\label{tab:criticality_mixed}
	\centering
	\begin{tabular}{lRRRRRR}
		\toprule
		\textbf{Planner}\hspace{-0.15cm} & \frac{m_\text{TTC}}{\si{s}} & \frac{p_\text{TTC}}{\si{\percent}} & \frac{m_\text{DRAC}}{\si{ms^{-2}}} & \frac{p_\text{DRAC}}{\si{\percent}} & \frac{m_\text{PET}}{\si{s}} & \frac{p_\text{PET}}{\si{\percent}} \\
		\midrule
		\textbf{Opt} & 3.4 & 4.3 & 1.3 & 12.6 & 1.4 & 26.7 \\
		\textbf{RL}  & 3.2 & 4.0 & 1.3 &  5.0 & 1.6 & 19.6 \\
		\textbf{NC}  & 3.2 & 4.8 & 1.3 &  8.4 & 1.6 & 18.5 \\
		\bottomrule
	\end{tabular}
\end{table}

\section{Real-World Experiments}
\label{sec:realeval}

To support our simulative results, we also conducted test drives in real-world traffic.
Although a similar comparative evaluation as in simulation is not possible due to fluctuating real-world traffic conditions, this shows that our cooperative maneuver planning system can be successfully deployed to a real-world setup of connected intelligent infrastructure and CAVs.

\subsection{Pilot Site Ulm-Lehr}
\label{ssec:pilotsite}

\begin{figure}
	\centering
	\includegraphics[width=\linewidth]{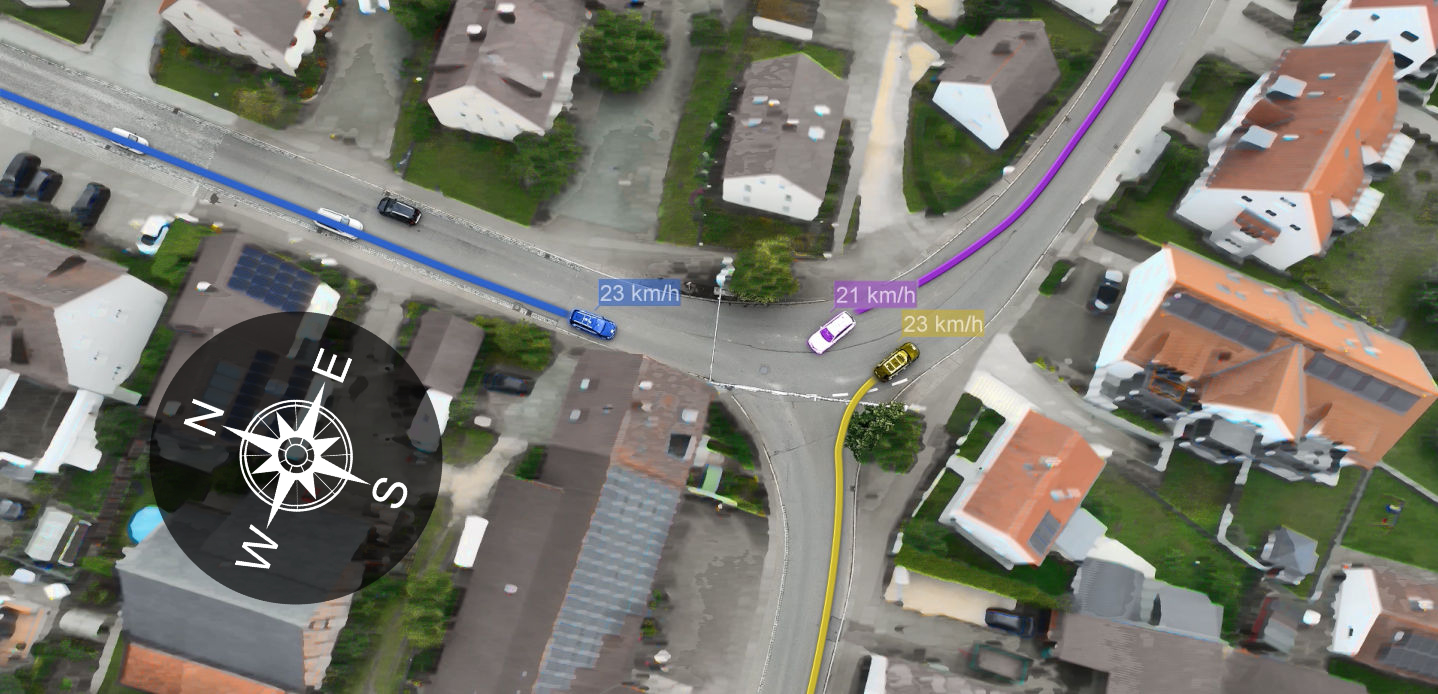}
	\caption{Aerial view of the pilot site in Ulm-Lehr, Germany~\cite{buchholz2021handling}, taken from the demonstration video~\cite{lukas2023video}.
		The unsignalized intersection is equipped with connected intelligent infrastructure.
		The picture shows a cooperative maneuver where the blue CAV $\nu_3$ on the bending main road yields to two other prioritized CAVs $\nu_1$ (purple) and $\nu_2$ (yellow) turning past each other.}
	\label{fig:realworld_aerial_view}
\end{figure}

The real-world experiments were conducted at a pilot site for connected automated driving in Ulm-Lehr, Germany~\cite{buchholz2021handling}.
An aerial view of the site is shown in Fig.~\ref{fig:realworld_aerial_view}.
The unsignalized intersection is equipped with multiple sensor processing units, each consisting of cameras, radar and lidar sensors as well as a computer for object detection.
A fusion and tracking module~\cite{Herrmann_2019_ITSC} on an edge server combines all object detections into the infrastructure EM, which is provided to the connected vehicles and the cooperative planning modules on the same edge server.
The server uses an AMD Ryzen 9 3950X CPU, similar to the simulation machine.
We employed three CAVs, one of which used the trajectory planning from~\cite{ruof_real-time_2023} while the other two implemented the concept from~\cite{voelz2019towards}.
The CAVs, infrastructure, and edge server are connected via a 5G SA cellular network with a communication latency of about {15\,ms}.

The scenario that was investigated in real-world experiments comprises the three CAVs approaching the intersection from each of its accesses, as depicted in Fig.~\ref{fig:intro}.
Vehicle~$\nu_3$ approaches the intersection on the main road coming from north and following the road.
According to ordinary priority rules, this vehicle has the right of way.
Vehicle~$\nu_1$ starts at the other main road access (east) and intends to turn left at the intersection.
This vehicle would have to yield to the oncoming traffic but not to any vehicle coming from the subordinate road.
With vehicle~$\nu_2$ approaching the intersection from the west on the subordinate road, it would have to yield to all cross traffic.
As it intends to turn right at the intersection, there is no conflict with~$\nu_1$ but only with~$\nu_3$.
Fig.~\ref{fig:intro} illustrates a typical cooperative maneuver.
If vehicle~$\nu_3$ waives its right of way and possibly slows down while approaching, it allows both other CAVs to cross the intersection virtually simultaneously.
Depending on the timing, vehicles~$\nu_1$ and~$\nu_2$ can simply traverse the intersection unimpeded.
Ideally, vehicle~$\nu_3$ should not be required to stop, but slow down smoothly and accelerate again once the intersection is cleared.

\subsection{Evaluation in Real-World}
\label{ssec:evalrealworld}

To obtain a reliable quantitative evaluation in real traffic, the initial scenario depicted in Fig.~\ref{fig:intro} was driven numerous times.
Thereof, five baseline runs have been performed by driving according to priority rules, twelve runs under the command of the optimization-based planner, and 20 runs employing the RL~planner.
The cooperative maneuvers have been planned reactively with respect to surrounding traffic, while the CAVs approach the intersection.
This manifests in alternative maneuvers being proposed if the original proposal turns out to be less effective or even infeasible.
One of the evaluation runs using the RL~planner was recorded on video for demonstration purposes and is available at~\cite{lukas2023video}.
The vehicle trajectories have been recorded and clipped to the common evaluation segment, which was also used for the simulative evaluation.
Because all experiments have been conducted in real traffic at a public intersection, each run had slightly different initial conditions.
While the safety drivers guided the test vehicles carefully to the engagement point, minor deviations in timing and initial speed remain. 
During all maneuver runs that are contained in the evaluation, the automated driving system was successfully engaged.

\begin{figure}
	\centering
	\includegraphics[width=0.9\linewidth]{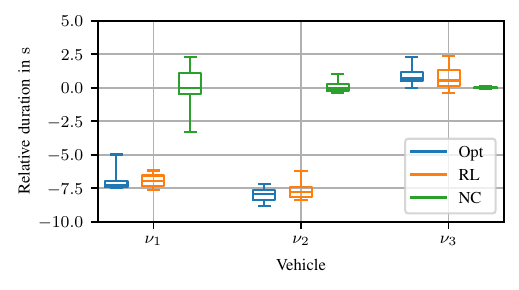}
	\caption{Real-world maneuver durations for each involved CAV relative to the median baseline duration.}
	\label{fig:realworld_relative_duration}
\end{figure}

One of the most expressive metrics for traffic efficiency at intersections is the delay induced by interaction with other vehicles.
In the present work, we consider the maneuver duration, i.e., the time it takes for a vehicle to cross the evaluation segment, relative to the median baseline duration.
Therefore, the median bars of the baseline distribution in the boxplot in Fig.~\ref{fig:realworld_relative_duration} are exactly at a duration of \SI{0}{\second}.
The outlier of vehicle~$\nu_1$ at around \SI{-3.5}{\second} can be explained by a deviation in initial timing.
In this run, the vehicle did not have to stop and its driving speed remained above \SI{3}{\meter\per\second}.
It can be observed that both cooperative planning approaches yield a significant reduction in maneuver duration for the vehicles~$\nu_1$ and~$\nu_2$, which cross the intersection simultaneously in the cooperative maneuver.
While the time gain for each of these vehicles is roughly \SI{7}{\second}, the vehicle~$\nu_3$ on the major road rarely looses more than \SI{2}{\second} during the approach and traversal of the intersection.
It can be concluded that both cooperative planners are able to identify traffic scenes in which the ordinary priority rules yield sub-optimal flow and propose a more efficient maneuver.

\begin{figure}
	\centering
	\includegraphics[width=0.9\linewidth]{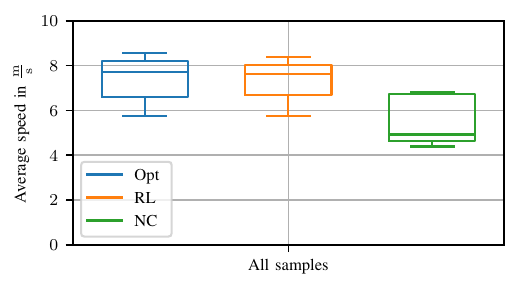}
	\caption{Real-world driving speed averaged over all CAVs when coordinated by one of the cooperative planners compared to baseline driving.}
	\label{fig:realworld_avg_speed}
\end{figure}

In addition, we present the average speed distribution over all vehicles in Fig.~\ref{fig:realworld_avg_speed}.
Both cooperative planners yield a mean driving speed of around \SI{7.5}{\meter\per\second} (\SI{27}{\kilo\meter\per\hour}) over the evaluated trajectory segment.
Considering the lane curvature on the confined intersection space, this indicates a swift intersection traversal without any stops.
In contrast, when driving according to priority rules, the average speed remains below \SI{5}{\meter\per\second}.
Undoubtedly, the yielding behavior of the vehicles~$\nu_1$ and~$\nu_2$ might lead to stops in this case, which causes a decrease in driving speed and efficiency.

The three real-world CAVs all use different motion planning algorithms that were adapted to demonstrate the cooperative use case and were not yet fully tuned to exactly meet the time constraints resulting from the maneuver planning.
Therefore, a criticality metric evaluation on the real-world test drives would not be of high significance, which should, of course, be addressed by further research.

\section{Conclusion}
\label{sec:conclusion}

In this paper, an optimization-based and an RL-based approach for cooperative maneuver planning in mixed traffic at urban intersections have been evaluated in a realistic simulation and in real-world experiments.
Especially in simulated fully automated traffic, cooperative maneuvers achieved a significant improvement of traffic efficiency.
Even in simulated mixed traffic with only 50\,\% CAV penetration, the planning approaches successfully performed cooperative maneuvers in about one third of the scenarios and increased the efficiency, despite not relying on any cooperation from HDVs.
Overall, the optimization-based planner yielded higher efficiency gains, while also causing a larger increase of criticality measures.
Experiments with three prototype CAVs in public traffic demonstrated the real-world applicability of both cooperative planners and showed a significant efficiency increase.
Future work will include the evaluation of further cooperative use cases in urban traffic involving human-driven connected vehicles and vulnerable road users.
Moreover, incorporating state uncertainty information from the environment model in the cooperative planning algorithms might be a promising avenue for future research.

\bibliographystyle{IEEEtranS}


\begin{IEEEbiography}[{\includegraphics[width=1in,height=1.25in,clip,keepaspectratio]{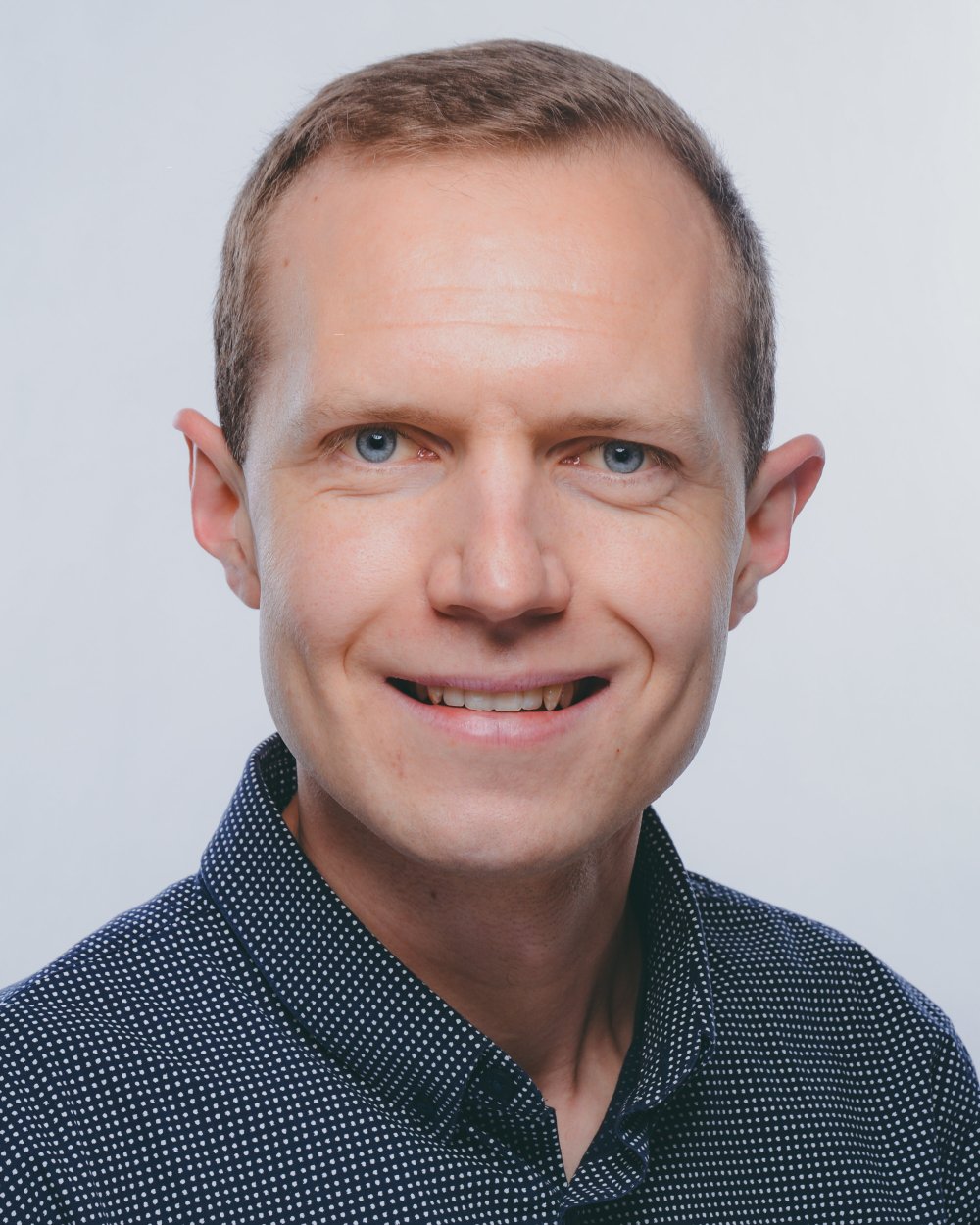}}]{Marvin Klimke } received his M.Sc. degree in Computer Engineering in 2021 from RWTH Aachen University and his Ph.D. degree in 2024 from Ulm University, Germany. He was a researcher with the Department of Corporate Research, Robert Bosch GmbH, Stuttgart, 70049, Germany, and the Institute of Measurement, Control, and Microtechnology, Ulm University, Ulm, 89081, Germany. His research interests include connected automated driving as well as behavior planning for automated driving by employing machine learning.
\end{IEEEbiography}

\begin{IEEEbiography}[{\includegraphics[width=1in,height=1.25in,clip,keepaspectratio]{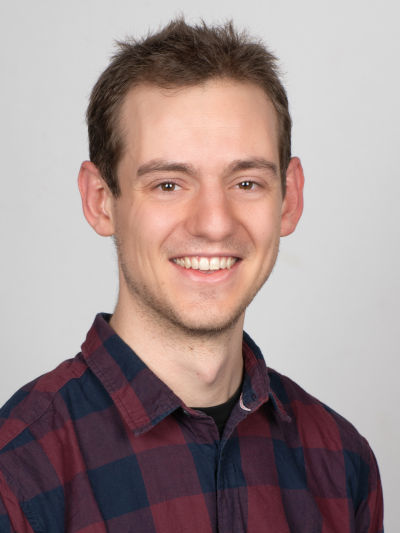}}]{Max Bastian Mertens } earned his M.Sc. degree in Communications and Computer Engineering at Ulm University in 2020. He is a researcher at the Institute of Measurement, Control, and Microtechnology, Ulm University, Ulm, 89081, Germany. His research interests include trajectory and cooperative maneuver planning as well as scene prediction in mixed traffic.
\end{IEEEbiography}

\begin{IEEEbiography}[{\includegraphics[width=1in,height=1.25in,clip,keepaspectratio]{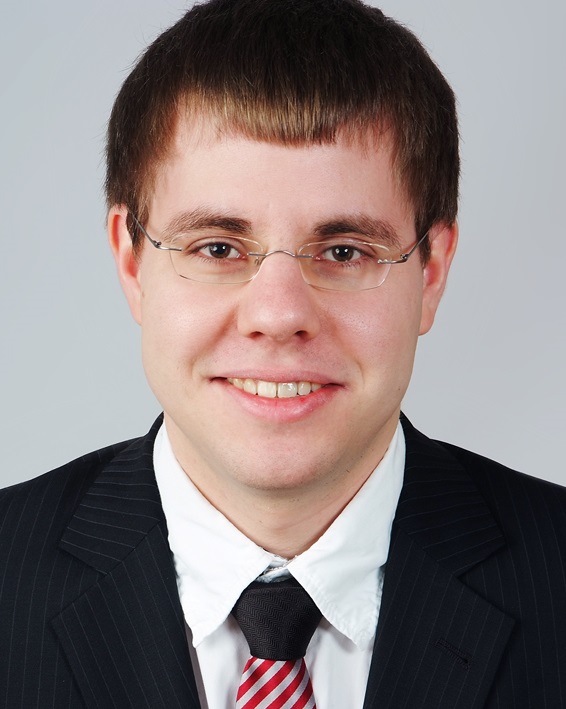}}]{Benjamin V\"olz } earned his diploma degree from the Faculty of Electrical and Computer Engineering, Dresden University of Technology, and his Ph.D. degree from the Department of Mechanical and Process Engineering, ETH Zurich. He is a research engineer at Robert Bosch, Stuttgart, 70049, Germany, focusing on planning for connected urban automated driving. His research interests include scene analysis, prediction, decision making, and planning for automated vehicles.
\end{IEEEbiography}

\begin{IEEEbiography}[{\includegraphics[width=1in,height=1.25in,clip,keepaspectratio]{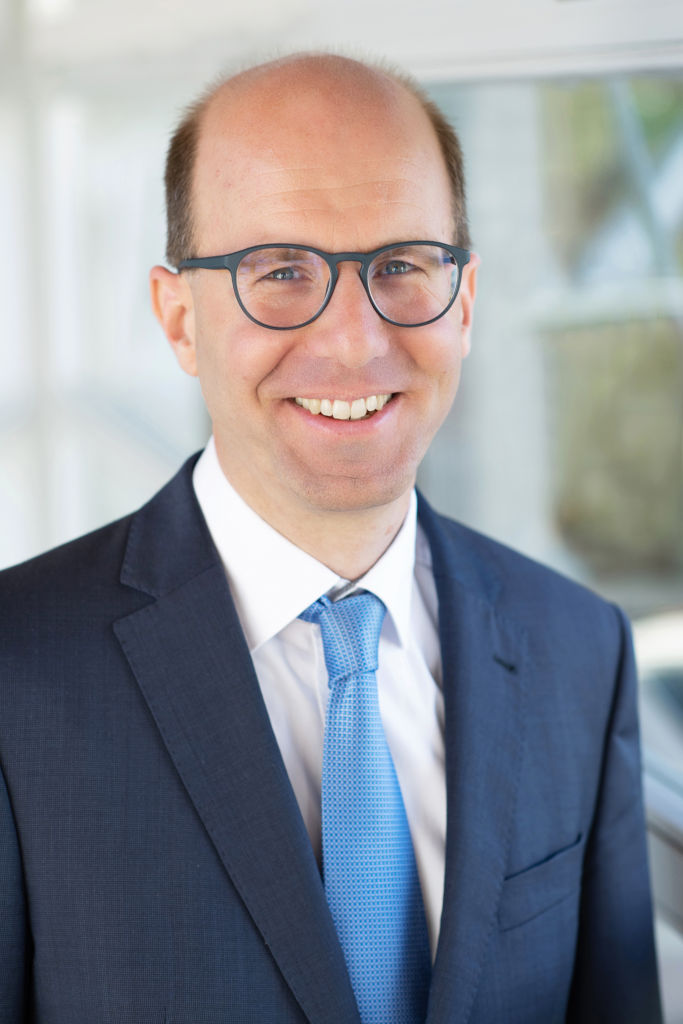}}]{Michael Buchholz } received his Diploma degree in Electrical Engineering and Information Technology as well as his Ph.D. from the Faculty of Electrical Engineering and Information Technology at University of Karlsruhe (TH)/Karlsruhe Institute of Technology, Germany.  He is a research group leader and lecturer at the Institute of Measurement, Control, and Microtechnology, Ulm University, Ulm, 89081, Germany, where he earned his \emph{Habilitation} (post-doctoral lecturing qualification) for Automation Technology in 2022 and the title of an \emph{apl. Professor} (adjunct professor) in 2025. His research interests comprise connected automated driving, electric mobility, modelling and control of mechatronic systems, and system identification.
\end{IEEEbiography}

\end{document}

%% file: img/fig1_intro.pdf_tex
\begingroup%
  \makeatletter%
  \providecommand\color[2][]{%
    \errmessage{(Inkscape) Color is used for the text in Inkscape, but the package 'color.sty' is not loaded}%
    \renewcommand\color[2][]{}%
  }%
  \providecommand\transparent[1]{%
    \errmessage{(Inkscape) Transparency is used (non-zero) for the text in Inkscape, but the package 'transparent.sty' is not loaded}%
    \renewcommand\transparent[1]{}%
  }%
  \providecommand\rotatebox[2]{#2}%
  \newcommand*\fsize{\dimexpr\f@size pt\relax}%
  \newcommand*\lineheight[1]{\fontsize{\fsize}{#1\fsize}\selectfont}%
  \ifx\svgwidth\undefined%
    \setlength{\unitlength}{252bp}%
    \ifx\svgscale\undefined%
      \relax%
    \else%
      \setlength{\unitlength}{\unitlength * \real{\svgscale}}%
    \fi%
  \else%
    \setlength{\unitlength}{\svgwidth}%
  \fi%
  \global\let\svgwidth\undefined%
  \global\let\svgscale\undefined%
  \makeatother%
  \begin{picture}(1,0.75)%
    \lineheight{1}%
    \setlength\tabcolsep{0pt}%
    \put(0,0){\includegraphics[width=\unitlength,page=1]{fig1_intro.pdf}}%
    \put(0.78591869,0.15361006){\makebox(0,0)[lt]{\lineheight{1.25}\smash{\begin{tabular}[t]{l}or\end{tabular}}}}%
    \put(0,0){\includegraphics[width=\unitlength,page=2]{fig1_intro.pdf}}%
    \put(0.65170395,0.2026611){\makebox(0,0)[lt]{\lineheight{1.25}\smash{\begin{tabular}[t]{l}predict\end{tabular}}}}%
    \put(0.49635742,0.03294522){\makebox(0,0)[lt]{\lineheight{1.25}\smash{\begin{tabular}[t]{l}$\nu_1$\end{tabular}}}}%
    \put(0.04631731,0.48972981){\makebox(0,0)[lt]{\lineheight{1.25}\smash{\begin{tabular}[t]{l}$\nu_2$\end{tabular}}}}%
    \put(0,0){\includegraphics[width=\unitlength,page=3]{fig1_intro.pdf}}%
    \put(0.65170395,0.15049265){\makebox(0,0)[lt]{\lineheight{1.25}\smash{\begin{tabular}[t]{l}select\end{tabular}}}}%
    \put(0.65170395,0.09970096){\makebox(0,0)[lt]{\lineheight{1.25}\smash{\begin{tabular}[t]{l}extract\end{tabular}}}}%
    \put(0.69855955,0.51608043){\makebox(0,0)[lt]{\lineheight{1.25}\smash{\begin{tabular}[t]{l}$\nu_3$\end{tabular}}}}%
    \put(0,0){\includegraphics[width=\unitlength,page=4]{fig1_intro.pdf}}%
    \put(0.63597856,0.33265727){\makebox(0,0)[lt]{\lineheight{1.25}\smash{\begin{tabular}[t]{l}Environment Model\\V2X Interfaces\end{tabular}}}}%
    \put(0,0){\includegraphics[width=\unitlength,page=5]{fig1_intro.pdf}}%
  \end{picture}%
\endgroup%

%% file: img/fig2_Aachen_Bendplatz_obs.pdf_tex
\begingroup%
  \makeatletter%
  \providecommand\color[2][]{%
    \errmessage{(Inkscape) Color is used for the text in Inkscape, but the package 'color.sty' is not loaded}%
    \renewcommand\color[2][]{}%
  }%
  \providecommand\transparent[1]{%
    \errmessage{(Inkscape) Transparency is used (non-zero) for the text in Inkscape, but the package 'transparent.sty' is not loaded}%
    \renewcommand\transparent[1]{}%
  }%
  \providecommand\rotatebox[2]{#2}%
  \newcommand*\fsize{\dimexpr\f@size pt\relax}%
  \newcommand*\lineheight[1]{\fontsize{\fsize}{#1\fsize}\selectfont}%
  \ifx\svgwidth\undefined%
    \setlength{\unitlength}{179.66803123bp}%
    \ifx\svgscale\undefined%
      \relax%
    \else%
      \setlength{\unitlength}{\unitlength * \real{\svgscale}}%
    \fi%
  \else%
    \setlength{\unitlength}{\svgwidth}%
  \fi%
  \global\let\svgwidth\undefined%
  \global\let\svgscale\undefined%
  \makeatother%
  \begin{picture}(1,0.46235042)%
    \lineheight{1}%
    \setlength\tabcolsep{0pt}%
    \put(0,0){\includegraphics[width=\unitlength,page=1]{fig2_Aachen_Bendplatz_obs.pdf}}%
    \put(0.91297345,0.26458324){\color[rgb]{0,0,0}\makebox(0,0)[lt]{\lineheight{1.25}\smash{\begin{tabular}[t]{l}$\veh_0$\end{tabular}}}}%
    \put(0.62826625,0.26249018){\color[rgb]{0,0,0}\makebox(0,0)[lt]{\lineheight{1.25}\smash{\begin{tabular}[t]{l}$\veh_1$\end{tabular}}}}%
    \put(0.23145312,0.39454774){\color[rgb]{0,0,0}\makebox(0,0)[lt]{\lineheight{1.25}\smash{\begin{tabular}[t]{l}$\veh_3$\end{tabular}}}}%
    \put(0.36965162,0.34149459){\color[rgb]{0,0,0}\makebox(0,0)[lt]{\lineheight{1.25}\smash{\begin{tabular}[t]{l}\SI{10}{m}\end{tabular}}}}%
    \put(0.04713528,0.25866595){\color[rgb]{0,0,0}\makebox(0,0)[lt]{\lineheight{1.25}\smash{\begin{tabular}[t]{l}$\veh_2$\end{tabular}}}}%
    \put(0.47675899,0.10703094){\color[rgb]{0,0,0}\makebox(0,0)[lt]{\lineheight{1.25}\smash{\begin{tabular}[t]{l}$d_{\text{target},1}$\end{tabular}}}}%
    \put(0.07552826,0.10764345){\color[rgb]{0,0,0}\makebox(0,0)[lt]{\lineheight{1.25}\smash{\begin{tabular}[t]{l}$d_{\text{stop},2}$\end{tabular}}}}%
    \put(0.17911906,0.32463099){\color[rgb]{0,0,0}\makebox(0,0)[lt]{\lineheight{1.25}\smash{\begin{tabular}[t]{l}$\Delta\psi_{3,10}$\end{tabular}}}}%
    \put(0.73193264,0.10680356){\color[rgb]{0,0,0}\makebox(0,0)[lt]{\lineheight{1.25}\smash{\begin{tabular}[t]{l}$d_{\text{lead},0}$\end{tabular}}}}%
    \put(0,0){\includegraphics[width=\unitlength,page=2]{fig2_Aachen_Bendplatz_obs.pdf}}%
  \end{picture}%
\endgroup%